# Broadening Ontologization Design: Embracing Data Pipeline Strategies

Chris Partridge
*BORO Solutions Ltd*
*University of Westminster*
London, UK
0000-0003-2631-1627

Andrew Mitchell
*BORO Solutions Ltd*
*University of Westminster*
London, UK
0000-0001-9131-722X

Sergio de Cesare
*Westminster Business School*
*University of Westminster*
London, UK
0000-0002-2559-0567

John Beverley
*Department of Philosophy*
*University at Buffalo*
Buffalo, USA
0000-0002-1118-1738

*Abstract*—Our aim in this paper is to outline how the design space for the ontologization process is broader than current practice would suggest. We point out that engineering processes as well as products need to be designed – and identify some components of the design. We investigate the possibility of designing a range of radically new practices implemented as data pipelines, providing examples of the new practices from our work over the last three decades with an outlier methodology, bCLEARer. We also suggest that setting an evolutionary context for ontologization helps one to better understand the nature of these new practices and provides the conceptual scaffolding that shapes fertile processes. Where this evolutionary perspective positions digitalization (the evolutionary emergence of computing technologies) as the latest step in a long evolutionary trail of information transitions. This reframes ontologization as a strategic tool for leveraging the emerging opportunities offered by digitalization.

## I. Introduction

In ontology engineering there is, in theory at the very least, a tight coupling between ontologies and ontologization, the process that produces them. Our aim in this paper is to suggest that the design space for the ontologization process is wider than a look at many of the current methodologies would indicate.

To illustrate this at a general level, we partition the space along two dimensions: the levels of generality and digitalization. Fig. 1 shows how this partitioned space is currently exploited – with a focus on the early stages of the process – exposing the areas that are not being exploited.

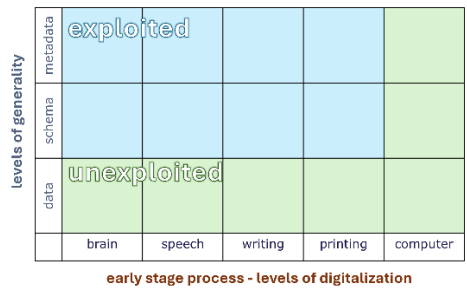

Fig. 1. Two-dimensional design space - exploitation

This suggests that the space is broader than current practices would indicate. That it is possible to open the space up to a range of potentially radical, new practices based upon data pipelines. We hope that by outlining some of these practices here we will make the case for broadening the design space and encourage the community to adopt a wider range of practices. In large part, the evidence for this is derived from our work over the last three decades with an outlier methodology, bCLEARer, which provides a useful example of these data pipeline practices.

We raise the engineering point that processes as well as products should be designed – and note a design poverty for formal ontologization processes relative to the final ontology products. We provide a factorization of the wider digitization process into which we believe formal ontologization fits. This separates computerization and ontologization concerns and we raise questions about how these components should be ordered.

It is recognized that for foundational issues, setting the right context can have a bigger impact on success than the quality of the problem-solving processes. This is the case for ontologization which needs contextual scaffolding to provide the perspective that enables one to better understand the scope and nature of these new practices. Specifically, that one should see ontologization as an essential part of a much wider more pervasive phenomena, the latest information evolutionary step – digitalization (the emergence of computing technologies). From a short-term perspective, this recapitulates the relatively recent steps of printing and writing. From a long-term perspective, this fits into an evolutionary trail of information transitions that spans life on earth. Within such perspectives, ontologization can be understood as a tool for exploiting the emerging opportunities offered by digitalization.

### A. Structure of the paper

In the next section, we provide a broad picture of the ontologization process and introduce two current mainstream ontologization methodologies to act as a baseline for comparisons. In the third section we introduce our example outlier data pipeline-based methodology, bCLEARer. In the fourth section we build the contextual scaffolding, firstly situating digitalization in an evolutionary perspective, then situating ontologization within digitalization. In the fifth section, we situate bCLEARer in this evolutionary perspective. In the sixth section, we illustrate from within the evolutionary perspective some of the outlier design choices that bCLEARer has made.

## II. A BROAD PICTURE OF THE ONTOLOGIZATION PROCESS

In ontology engineering one would expect a tight coupling between ontologies and ontologization, the process that produces them. One can characterize this as a product-process distinction, a fundamental concept in traditional engineering. The engineering mindset differentiates the product (the output) and the process (the method or system used to produce the output) and expects both to be engineered. And, as part of the engineering, to both be quality managed, hence quality assurance (process) and quality control (product).

### A. Top-level ontologies and ontologization

In top-level ontology (engineering) work, the process is often an invisible relation of the product. An example of this is provided by the main standard, ISO 21838-1:2021 – Information technology: Top-level ontologies (TLO) – Requirements [1]. While this references the ontology of processes, the standard makes no mention of the ontologization process itself. Hence, unsurprisingly, the standards based upon it do not mention the ontologization process either.

Some top-level ontologies have associated documentation for the ontologization process. The top-level Descriptive Ontology for Linguistic and Cognitive Engineering (DOLCE) has a related analysis tool OntoClean [2], but this falls far short of an ontologization process. The Basic Formal Ontology (BFO) has a book [3] on the ontologization process that assumes the BFO top-level ontology. We look at this text in more detail below. The BORO Foundational Ontology has a closely intertwined bCLEARer ontologization process described in [4] and [5]. There are a variety of domain level ontologization processes that we discuss later in this paper.

### B. The case for engineering the ontologization process

The importance of engineering the process is reflected in the often-quoted dictum that: the quality of the process determines the quality of the product.

For a historical background to this, from the wider history of innovation, see Mokyr's *The Past and the Future of Innovation* [6] or his *A Culture of Growth* [7]. He argues that history shows that technological progress cannot rely on artisanal skills alone, it needs to be supplemented with formal and systematic (that is, engineered) knowledge.

Within engineering, this idea was explored, analyzed and championed in manufacturing in the second half of the 20$^{th}$ century by quality management pioneers such as W. Edwards Deming [8] and Joseph Juran [9]. This led to a rich variety of designs including movements such as Total Quality Management and its successors Lean Manufacturing, and Six Sigma. These developed a fertile range of ways of managing manufacturing processes. An example is the Plan-Do-Check-Act (PDCA) Cycle used to design, implement, and refine processes on a small scale – which fits well with the Kaizen philosophy of continuous improvement, where processes are regularly reviewed and improved incrementally. This has spread to some other domains. For example, one can see Kaizen-like principles being used in the Agile software development methodology.

Within the ontology engineering community, one does not find a comparatively rich selection of designs and range of ways of managing the ontologization processes – a kind of process design poverty. This is despite a few interesting innovative examples such as the ROBOT tool (https://robot.obolibrary.org/extract.html). Especially for top-level ontologies, there appears to be more 'theory' for, and so more attention on, the design of the final product – 'the 'ontology' – than the process – 'ontologization' – that produces it.

From an engineering perspective, this imbalance looks unhealthy. One could argue that this poverty arises from the process being relatively new and under-researched, unlike, for example, top-level ontology which can build upon a rich heritage.

#### 1) Process design poverty in logic

Interestingly, a similar poverty of design process has been pointed out in logic, which is a key part of the last stages of the 'ontologization' process. Novaes [10] makes a product-process distinction for logic, distinguishing the formal product from its formalization process, noting an almost exclusive focus on the former in contemporary logic:

*"As a discipline, logic is arguably constituted of two main sub-projects: formal theories of argument validity on the basis of a small number of patterns, and theories of how to reduce the multiplicity of arguments in non-logical, informal contexts to the small number of patterns whose validity is systematically studied (i.e. theories of formalization). Regrettably, we now tend to view logic 'proper' exclusively as what falls under the first sub-project, to the neglect of the second, equally important sub-project."*

She discusses two historical theories of argument formalization, from Aristotle and Medieval Logic that have more balance. Both "illustrate this two-fold nature of logic, containing in particular illuminating reflections on how to formalize arguments (i.e. the second sub-project)." She suggests reflecting on these should lead to a broader conceptualization of what it means to formalize.

Given how much the ontology (engineering) product builds on formal logic – inheriting many of its (cultural) practices – this may contribute to the poverty in ontology engineering. This suggests that developments in the formalization process could be recruited by and enrich ontologization's approach to formalization.

### C. Comparing different ontologization processes

In this section, we briefly look at some current mainstream methodologies that guide the ontologization process to provide a basis for comparison with the data pipeline approach exemplified by bCLEARer. This gives us a rough benchmark on common practices. A caveat: we do not claim that this selection reflects all the work that is happening in this area. Rather, we are aiming for examples that lend themselves to our broad comparison.

In this section we restrict ourselves to ontologization to help make a clear comparison. This is even though, as we touch upon later from a bCLEARer data pipeline perspective, there are interesting features in the methodologies guiding the processes in other software related domains, such as:

- Waterfall model: clear top-down separation of concerns.
- Agile: flexible, responsive, efficient, iteration
- DevOps (and DataOps): automation into a data pipeline to improve and shorten life cycles.

There is a reasonably rich literature on ontology methodologies, including [11], [12], [13], [14], [15], [16], [17]. We roughly divide these into two broad camps, which we have colloquially labelled: 'Ask-an-Expert' (AaE) and 'Top-Down-Classification' (TDC). We have selected a representative document for each camp: For AaE, OntoCommons report D.4.2 [18] and for TDC, *Building ontologies with Basic Formal Ontology* [3].

One aspect of these methodologies we inspect is the information pathway they create, the flow or movement of information through the stages of the overall process. We specifically explore how this interacts with the two dimensions of the design space. Firstly, the level of generality dimension which, for ease of understanding, we introduce from a data perspective as the metadata, schema and data levels. This is a simplification as it is about the syntax of the implementation, whereas generality is also a semantic matter. However, there is a good enough rough match between syntax and semantics here to make the substitution fair for our broad classification. Secondly, the levels on the journey to digitalization dimension [19], [20]. This looks at the evolutionary steps on the journey to digitalization. Very broadly a journey that goes from brains to speech to writing to printing and then computing. We call this the levels of digitalization. The results of the inspection are in the earlier Fig. 1.

*1) The 'Ask-an-Expert' approach*

We selected the OntoCommons report D.4.2 [18] as our basis for AaE. It suits our purposes as it not only describes its own approach (the LOT methodology) but documents other similar approaches (including Grüninger & Fox [13], METHONTOLOGY, On-To-Knowledge, DILIGENT, NeOn, RapidOWL, SAMOD and AMOD). Together these provide many good examples of the 'Ask-an-Expert' (AaE) approach, which has its roots in Artificial Intelligence (AI) and knowledge representation.

This process is largely a rationalist armchair exercise – in the sense that there is little empirical content. The input for the process is domain experts – as this quote illustrates:

*"The goal of the ontology implementation activity is to build the ontology using a formal language, based on the ontological requirements identified by the domain experts." [13, p. 28]*

Across all the approaches reviewed, there is a similar information pathway from a level of digitalization perspective. In the early stages there is an underlying focus on natural language (from a levels of digitalization view, speech), sometimes organized into (natural language) competency questions [13] – as this quote illustrates:

*"If domain experts have no knowledge about ontology data generation and querying, we recommend writing the requirements in the form of natural language sentences." [13, p. 22]*

The methodology's input to the information pathway is the brains of experts via speech into documented (unstructured) natural language. Then the methodology broadly separates concerns [21]: separating the confirmation of content from its formalization – and chooses to address the first concern before the second.

We will revisit this point later, but it is important to note that this separation and ordering choice assumes that reaching content agreement prior to the formalization process won't negatively impact the final product. In this design architecture, the first stage is a confirmation of content which uses mostly (unstructured) natural language which is organized and agreed as a statement of the requirements of the ontology. The second stage takes the natural language and formalizes them.

The early pathway is not always or entirely natural language, as there is a mention of the possibility of using more structured information in the shape of a "tabular technique" using "3 types of tables: Concepts, Relations, Attributes". Formalization (structured information) only really enters the process in the later stages of the pathway in ontology implementation, after the requirements (expressed in natural language) are collected.

The paper notes that there is optionally a conceptualization stage, where an interim concept model based upon the requirements may be built. Interestingly, it suggests that "*diagraming tools such as MS Visio or draw.io, as well as non-digital tools as pen and paper or a blackboard*" may be used to build this.

*2) The 'Top-Down-Classification' approach*

We take *Building ontologies with Basic Formal Ontology* [3] as the baseline for the 'Top-Down-Classification' (TDC) approach. This provides a clear example with a concise summary of how it aims to construct an ontology (this shows why it deserves the top-down-classification nickname). This process is also largely a rationalist armchair exercise, one that has roots in biological classification and philosophy. It is a common approach to developing top-level ontologies in Information Systems (IS).

*"**Overview of the Domain Ontology Design Process***
*Ontology is a top-down approach to the problem of electronically managing scientific information. This means that the ontologist begins with theoretical considerations of a very general nature on the basis of the assumption that keeping track of more specific information (for example, about specific organs, genes, or diseases) requires getting the very general scientific framework underlying this information right, and doing so in a systematic and coherent fashion. It is only when this has been done that the detailed terminological content of a specific science such as cell biology or immunology can be encoded in such a way as to ensure widespread accessibility and usability." [3, p. 49]*

This informal view is then structured into a step-by-step process in a table – see below.

*Table 3.1*
*An outline of the steps to be followed in designing a domain*

*ontology*
*1. Demarcate the subject matter of the ontology.*
*2. Gather information: identify the general terms used in existing ontologies and in standard textbooks; analyze to remove redundancies.*
*3. Order these terms in a hierarchy of the more and less general ones.*
*4. Regiment the result in order to ensure:*
 *a. logical, philosophical, and scientific coherence,*
 *b. coherence and compatibility with neighboring ontologies, and*
 *c. human understandability, especially through the formulation of human-readable definitions.*
*5. Formalize the regimented representational artifact in a computer usable language in such a way that the result can be implemented in some computable framework.*
[3, p. 50]

From this table, we can pull out a level of digitalization perspective along the information pathway. The first four stages work with unstructured natural language. Though the terms 'order' and 'regiment', at steps 3 and 4, suggest some structure in the information, it is only at step 5 that the information is formalized, and so fully structured data enters the process. So here as well, the information pathway to the ontology starts with brains then via speech or directly into text is documented (unstructured) natural language.

In the process, there is a similar reliance upon human experts to justify choices, see:

*"The terms in an ontology are the linguistic expressions used in the ontology to represent the world, and drawn as nearly as possible from the standard terminologies used by human experts in the corresponding discipline." [3, p. 5]*

As an aside, it is often not recognized that the terms themselves, as inscriptions or utterances, are also elements of the domain that can usefully be represented in the ontology

*3) Process Comparison*

One can make a rough assessment of the engineering maturity of these methodologies. As the quotes above hint at, they are currently collections of "ad hoc rules" with simple heuristics. There is no background context to act as a foundation to guide the engineering of the process design – certainly no common context. Hence, they are, from an engineering design perspective, at an early stage of development. There is still plenty of scope for them to undergo the kind of serious engineering re-design Deming and Juran undertook for manufacturing.

Both approaches have several features in common, ones that differentiate them from data pipeline approaches, such as bCLEARer. From the perspective of digitalization, in both cases, their early processes for establishing the base ontology focus their efforts on working with unstructured (pre-digitalization) natural language related to human understandability, with less focus on machine understandability. In both approaches the ontologization happens before the formalization. The (unstructured) ontological information is captured and regimented in natural language first and then formalized.

One can broadly divide information into levels of generality. From a syntactic 'data' perspective, these are the natural levels: metadata, schema and data. For our purposes here, these are a good enough rough proxy for the semantic levels; top, the most general, middle and the most specific bottom level – typically particulars.

We can use this perspective to show how the two approaches showcased differ and agree. They differ in their approach to the metadata level. The 'top-down-classification' approach works by framing the middle level in terms of the relevant top-level structure – so the middle schema level is framed by this metadata. The 'ask-an-expert' approach works explicitly at the schema level – focusing on the domain. In principle, there is no reason why the 'ask-an-expert' approach could not start with the metadata level, or the 'top-down-classification' approach could not ignore the metadata level and work at the schema level. The two approaches agree on their approach to the bottom data level. They both ignore it (a significant omission, we return to below).

One can relate the design of the digitalization and generality features. If one chooses to design a process to work with humans and unstructured information, one needs to recognize that one is building in scaling constraints that block the processing of large amounts of information. One way around this is, of course, to work with the data level indirectly through the schema level. Data pipeline approaches take a different route and aim to automate the process so removing these scaling constraints.

### III.  bCLEARer – History

In this section, we provide more context with a brief background history of the data pipeline approach, bCLEARer and its place in BORO (an acronym for '*Business Object Reference Ontology*'). BORO's development and deployment started in the late 1980s. This early work is described in *Business Objects* [4]. BORO's focus was then, and is now, on enterprise modelling; more specifically, it aims to provide the tools to salvage and reuse the semantics from a range of enterprise systems building a single ontology with a common foundation in a consistent and coherent manner.

BORO was originally developed to address a particular need for a solid legacy re-engineering process. This naturally led to the development of a methodology for re-engineering existing information systems, currently named bCLEARer – where the capital letters are an acronym for Collect, Load, Evolve, Assimilate and Reuse. This was co-developed with a closely intertwined top-level ontology (the BORO Foundational Ontology). Hence, the term BORO on its own can refer to either of, or both, the mining methodology and the ontology.

Our focus here is on the bCLEARer methodology which is used to systematically unearth reusable and generalized ontological business patterns from existing data. Most of these patterns were developed for enterprises and successfully applied in commercial projects within the financial, defense, and energy industries.

bCLEARer has evolved organically over the last three decades both in response to the evolutionary pressures of experience as well as exploiting the opportunities provided by evolving digital technology. An early version of the methodology is described in [4] – with a detailed description in

Part 6. At the time this was developed, the late 80s and early 90s, the technology support was immature, so while the core process was systematized it was not fully automated. Over the last decade, as appropriate technology has emerged, the core process has been fully automated into a data pipeline. Later versions of this are described in several places, including [22]. There are also open-source examples on GitHub (https://github.com/boro-alpha).

bCLEARer (and its associated top-level ontology) have, over the years, been configured to exploit a variety of situations ranging from its original legacy system migration to application migration to developing requirements and quality controlling existing systems. A common feature of all these projects has been the initial collection of one or more datasets (where this may include both structured and unstructured data – though structured data is preferred) and its regimented evolution to a more digitally aligned state.

IV. SITUATING DIGITALIZATION AS INFORMATION EVOLUTION

We have established that (engineers have learnt that) the quality of the final product depends upon the engineering quality of the design of the process. We have also suggested that the mainstream ontologization processes are very lightly engineered with a weak background context. This indicates that there is an opportunity to develop a more engineered ontologization process. What is less clear is what form this engineering should take.

Over the last three decades, the evolution of bCLEARer's practices was initially driven by experience. This pragmatic, experiential approach is supported by many including Aristotle [23] who said, "for the things we have to learn before we can do them, we learn by doing them". Reflecting upon the process has always been a central part of the practice. However, in the last decade, as the practice has matured, questions about the broader context have naturally arisen and this has led to a much better understanding of how the practice should be engineered.

This better understanding has merged in large part from a recognition that for foundational issues, setting the right context can have a bigger impact on improvements than the quality of the problem-solving processes. This is not a new idea; it is already established in many fields. In the context of professional practice, Donald A. Schön's "The Reflective Practitioner," [24] raised the concept of problem setting as a crucial part of having a successful outcome. In design thinking, David Kelley emphasizes the importance of problem framing in his book "The Art of Innovation" [25] explaining that reframing the problem often opens the door to more creative and effective solutions. In systems thinking, Peter Senge's "The Fifth Discipline" [26] differentiates between problem identification and problem solving. For him effective problem solving involves identifying leverage points – places within a system where a small change can lead to significant, long-term improvements.

In each of these cases, the initial stage involves developing a clear understanding of the underlying causes and interconnections of the entire complex system. Problem solving then involves developing interventions that address the root causes.

In the case of ontologization there is a ready-made context that can provide the perspective needed – this is evolutionary theory. If one positions ontologization as an essential part of a much wider more pervasive phenomena, the latest information evolutionary step – digitalization (the emergence of computing technologies) then a new picture emerges. From a short-term perspective, this recapitulates the relatively recent steps of printing and writing. From a long-term perspective, this fits into an evolutionary trail of information transitions that spans life on earth. Within such perspectives, ontologization can be understood as a tool for exploiting the emerging opportunities offered by digitalization.

We can then position the bCLEARer practices into the overall evolutionary context. This leads, in turn, to a clearer picture of the specific evolutionary pressures within the practice. From the start, bCLEARer has been framed in terms of information engineering, evolution and revolution. The narrative in [4] was the evolution of information paradigms. For much of its early life the focus of bCLEARer's evolution has been pragmatic practice adapting to the evolutionary pressures presented by actual ontologization with only a modicum of reflection upon the nature of the process. In the last decade or so there has been more reflection on what these practices might reveal. We are reaching a conclusion that the best way to understand the design of the process is to make the originally evolutionary framing much richer – to firstly more clearly frame the process as digitalization and secondly to show the digitalization as part of a wider trend that frames evolution in terms of information.

In the next section, we set up a broad picture of this wider trend of evolving information. In the section after that we situate digitalization as one transition in that evolution.

*1) Situating digitalization as information innovation*

Digitalization is an information transition – and it turns out information transition is a ubiquitous pattern which can be used to frame the whole of macro-evolution. This provides a reassuringly broad context where digitalization is the latest in a long history of information transitions.

Maynard Smith and Szathmáry [27], [28], [29] suggest that macro-evolution can be characterized as a series of information transitions and that one can frame the whole of macro-evolution in these terms:

*"… that evolution depends on changes in the information that is passed between generations, and that there have been 'major transitions' in the way that information is stored and transmitted, starting with the origin of the first replicating molecules and ending with the origin of language."*

And these changes in information transmission, the passing of *"information … between generations"*, are central to evolution. Maynard Smith and Szathmáry provide a table of seven major transitions, where the third is the "genetic code" and the seventh is "language". Each transition not only transforms life but also transforms the way life evolves – and so, in a sense, evolution evolves through transformations in information transmission. They also suggest that each of these transitions has accelerated and expanded evolution enabling more complex entities to emerge quicker.

Jablonka and Lamb [30] expanded this framework saying:

*"… we argue that information transmitted by non-genetic means has played a key role in the major transitions, and that new and modified ways of transmitting non-DNA information resulted from them."*

More specifically, they argue that:

*"The evolution of a nervous system not only changed the way that information was transmitted between cells and profoundly altered the nature of the individuals in which it was present, it also led to a new type of heredity—social and cultural heredity—based on the transmission of behaviorally acquired information."*

This new type of heredity enables even faster, more flexible, more complex evolution. One key feature is that the social and cultural heredity is not (like genetic heredity) necessarily dependent upon (genetic) life cycles, so it can give rise to adaptations which easily spread through a population within a life cycle. This social and cultural adaptation is orders of magnitude faster than genetic evolution and, particularly in changing environments, faster adaptive evolution is more successful. Human culture is a good example of this.

In *Evolution in four dimensions: genetic, epigenetic, behavioral, and symbolic variation in the history of life* [31], Jablonka and Lamb, describe in Chapter 9 – *Lamarckism Evolving: The Evolution of the Educated Guess* how new types of non-genetic heredity – behavioral and symbolic inheritance – enable a new directed evolution. They label this, understandably, Lamarckian.

*"… the variation on which natural selection acts is not always random in origin or blind to function: new heritable variation can arise in response to the conditions of life. Variation is often targeted, in the sense that it preferentially affects functions or activities that can make organisms better adapted to the environment in which they live. Variation is also constructed, in the sense that, whatever their origin, which variants are inherited and what final form they assume depend on various "filtering" and "editing" processes that occur before and during transmission."*

*a) Evolutionary transitions in information transmission*

This Lamarckian 'targeting' and 'constructing' enables further evolutionary transitions in information transmission. Obvious examples from symbolic evolution are the emergence of writing and printing information technologies. These involve fundamental "changes in the way information is stored and transmitted" where writing involved changes in structure and printing changes in economics. These both clearly led to further innovations in information evolution. Ironically, this reveals CRISPR technology, where DNA is selectively modified, as a Lamarckian genetic evolution.

While the transitions clearly involve the use of new technology, closer examination (see, for example, Ong [32] and Olson [33]) reveals they depended upon the coevolution of human behavior and external information technologies. Where both the emergence and exploitation of the technology depends upon intertwined coevolution with human behavior.

There is a similar intertwined evolution of behavior and technology so far in the emergence of computing technology. This current transition is often broadly called, in the context of enterprise processes, 'digitalization' – which includes 'digitization', the process of converting information, data, or physical objects into a digital format, readable by computers.

*b) Domestication as an example of co-evolution*

A more familiar example may help us to appreciate the nature of coevolution – the domestication of plants and animals. This has been studied as a distinctive coevolutionary relationship between domesticator and domesticate in a range of research [34], [35]. In this domain, it is easy to see that both parties (the domesticators and domesticates) coevolve in the sense of contributing to the relationship. Zeder [34, p. 3191] describes domestication as a:

*"… relationship in which one organism assumes a significant degree of influence over the reproduction and care of another organism in order to secure a more predictable supply of a resource of interest."*

Our relationship with the new digital forms of information technology can be described in a similar way. One where we domesticate our computer systems controlling their breeding.

*c) A new 'digital' form of information transmission*

In *The Selfish Gene*, Dawkins [36] introduced an idea. He distinguished between genes as "replicators" that pass on copies of themselves through generations and organisms as "vehicles" or "survival machines" constructed by genes to survive in the environment and so ensure their continued replication.

One could adopt a 'promiscuous ontology', one that regards computer systems as a form of life subject to evolution [37], [38], [39]. If so, then computers can easily be seen as "vehicles" for the information they carry and replicate as well as things we domesticate and breed. Furthermore, the emergence of these individuals then passes the Smith and Szathmáry test in so far as it radically 'changes the way information is stored and transmitted' directly between these individuals – and with humans.

One could be less adventurous and instead see the computer systems as an extension of humans [40]. In this view, computer systems are replicators rather than vehicles – they are part of the apparatus transmitting information rather than "survival machines" in a computer ecosystem. Even on this view, they pass the Smith and Szathmáry test. So, either way, one can see them as providing an opportunity for an information transition [20].

*2) Narrower context – Lamarckian choices for co-evolution*

If, as looks likely, history repeats itself and this digital transition follows the pattern of most previous transitions, then it is likely to involve the coevolution of human behavior and digital technology. The energy enterprises currently devote to their digitalization efforts show an intuitive understanding that some kind of directed effort needs to be made. From our evolutionary perspective, we can see this as our culture starting to co-evolve with the new technology – looking to construct the Lamarckian variations that will exploit this opportunity. What is less clear is which Lamarckian constructed variations are likely to lead to significant success – in the language of evolution to be able to exploit the natural selection pressures well.

Maybe history has a clue. A common historical narrative is that technology drives change, that it is the emergence of a new technology that initiates the associated cultural change. Careful study reveals a more interrelated pattern of co-evolution between technology and culture. Where cultural change often prepares the ground for technological innovation and then feeds off it and feeds further innovation. Olson [33] provides a relevant example, explaining how cultural developments in Western Europe from the 9th century onwards played a key role in the invention of moveable type in the 15th century. This technological innovation then laid the ground for developments in Western science in the 16th and 17th centuries. So maybe the coevolution of culture and digital technology evolution can give us some clues on where to target Lamarckian variations.

It is well-accepted that work in logic and mathematics laid the foundation for computing. If we look at the culture of this work, then we can see some trends that help us to target variations to help the co-evolution. Well before the introduction of digital computers, Frege [41] uses the analogy of a microscope and the eye to explain how his formal language compares with ordinary informal language, noting that it provided a superior sharpness of resolution. Carnap [42] talks about 'rational reconstruction' (rationale Nachkonstrucktion). Quine [43] says that one doesn't merely clarify commitments that are already implicit in unregimented language; rather that one often creates new commitments by regimenting. Quine notes that paying attention to the ontological commitment often leads to radical 'foreign' differences [44, pp. 9–10]: "Ontological concern is not a correction of lay thought and practice; it is foreign to the lay culture, though an outgrowth of it" Adding "There is room for choice, and one chooses with a view to simplicity in one's overall system of the world." Lewis [45, pp. 133–5] following the theme of 'outgrowth', argues that the differences are a result of taking the lay common sense seriously, by trying to make it simpler and consistent.

*3) The ontologization value chain*

The bCLEARer methodology is an example of data pipeline approach, which is the topic of this paper. It has, through experience refined these historical intuitions, developing a view of the digitalization process as a network of transformation processes in a data pipeline. One way to characterize this is as a value chain, where each transformation adds value. We briefly outline what a value chain is below and then describe the factorization of digitalization into component transformations.

*a) Recruiting the value chain view*

In manufacturing, Porter's value chain [46] provides a useful tool for broadly characterizing processes as a system of transformations. The system is provided with inputs which feed into a network of transformation processes. This network feeds into the output. The characterization is recursive. Each transformation process can be seen as a sub-system with its own value chain.

We recruited a lightweight version of this tool to characterize ontologization. Under this view, at the broadest level, the ontologization process starts with pre-ontologization information and is transformed using an ontologization process into an ontology. It adds value by transforming the pre-ontologization information into a formal ontology. This reveals an information pathway that starts with the pre-ontologization inputs, undergoes transformations and is output as a formal ontology. Different methodologies have different intermediate transformations and so different information pathways. While the inputs and outputs remain similar, the value chain transformations differ.

*b) Factoring digitalization into \*computerized and \*ontologized*

We firstly factorize digitalization into two types of digital transitions that it has found useful to target (and construct). These are *computerization and *ontologization. We use the '*' prefix convention to indicate our specialized use of the term and differentiate it from the many other senses in which it is used.

*Computerization is the process of converting relatively unstructured information into formally structured data. It implies something more than the digitization mentioned earlier, which just aims at bare computer readability. Formally structured data refers to information that is organized into a highly defined and predictable form, typically within a fixed schema or format. So, a scan of an engineering drawing in, say, PDF format would be digitized but not *computerized, as there is no direct way for the computer to read the components of the drawing. Whereas an engineering drawing in a CAD format, such as native DWG, would be *computerized, as the information in the drawing is explicit in its structure and can be read directly by a computer. *Computerization is intended to be a pragmatic distinction and while there are borderline cases, there are also cases that clearly fall into the pre-*computerization and *computerization camps.

*Ontologization is the process of converting relatively semantically unorganized information into information organized into a common ontological structure. Typically, the information is used in a domain, and there is a level of semantic precision needed for it to be fit for purpose. The *ontologization organizes the information into a common ontological structure that is sufficiently fine-grained to capture the requisite semantic precision.

One way of characterizing *ontologization is that it develops an explicit picture of ontological commitment [47], [48]. There is a long tradition of seeing this process as a transformation that reveals a deeper structure.

Currently, most information systems being digitized have no precise explicit ontological commitment. So, in practice, the *ontologization is often a regimentation [43] and rational reconstruction [42] of what the ontological commitment would be given some preferred top-level ontology. In bCLEARer's case, the top-level ontology is the BORO Foundational Ontology [5].

*c) \*Computerization – transitioning from implicit to explicit formal structure*

The bCLEARer methodology has further identified a factorization of the *computerization transition into two sub-transitions: surface-*computerization and deep-*computerization, corresponding to two levels of *computerization.

The process of transforming unstructured pre-*computerized information, giving it a highly defined and predictable form is sufficient to make it surface-*computerized. Much data in data stores is in this state today. It may, and often in practice does, have significant implicit formal structure. In some cases, making the structure implicit may be deliberate, as part of the process of improving the performance of a system. For our purposes we want to make the structure explicit to facilitate the *ontologization. We do this in the process of deep-*computerization.

Uncovering the underlying implicit form of surface-*computerized information requires a degree of ethnographic hermeneutics – one needs to be able to interpret, to understand, its implicit structure from its perspective. The deep-*computerization transformation aims to expose this and, as far as feasible, make the underlying infrastructure transparently clear.

A simple example of surface-*computerized information would be SQL table schemas and their associated data without the foreign keys noted. This meets the criteria for being *computerized – it has a fixed format. However, when the deep-*computerization adds the foreign keys, one can appreciate that the pre-deep-*computerization information had implicit structure that was not explicitly visible to a computer reading the data. In other words, it was only surface-*computerized.

There are existing software techniques that work in this space, that one can build upon. These include refactoring [49] and clean coding [50]. Both are bodies of practices for restructuring existing code, altering its internal structure to improve it, without changing its external behavior. The restructuring can be recruited to reveal the deeper structure.

From an ethnographical perspective, deep-*computerization (and maybe surface-*computerization too) is a kind of 'surfacing'. As Star notes in *The Ethnography of Infrastructure* [51] the details are technical and "excursions into this aspect of information infrastructure can be stiflingly boring". This means that large parts of the infrastructure are often invisible, in the sense that one doesn't pay attention to them. So, one of the challenges is training oneself to see, and so surface, the invisible structure – a practice with similarities to Bowker's [52] "infra-structural inversion", which foregrounds the backstage operational elements.

As with the other factorizations, this is intended to be a pragmatic distinction where most cases clearly fall into one or other camp – but with some borderline cases. One common borderline case is data cleansing. This includes technical matters such as resolving encoding issues and the treatment of whitespaces (which we find are both still common) as well as keying and spelling errors. While these might degrade the quality of the surface-*computerized information, they do not seem sufficiently grave to undermine its *computerized status. And fixing them does not obviously qualify as immediately revealing implicit structure – though if they are not fixed, they can hide structure. For pragmatic reasons, we take fixing these to be part of the deep-*computerization process.

*d) Inter-process dependency*

Obviously, there is an order to the surface- and deep-*computerization process. One surface-*computerizes information before *deep-computerizing it. Theoretically, at least, the *computerization and *ontologization processes would seem to be sufficiently independent that one could undertake either one without the other – implying that there is a choice in which to do before the other.

However, the bCLEARer experience is that there are strong pragmatic reasons for undertaking the full *computerized transition before undertaking the *ontologization transition [48], [20]. Our experience has been that the formalization process inherent in *computerization is best done with raw unaltered data, straight from the operational 'wild'. This is because we found that in cases where the *ontologization process was carried out on pre-*computerization information, it often obfuscated structure that *computerization needed – making the overall process significantly harder. Hence, in bCLEARer we see *ontologization as primarily a process for refining already *computerized data.

## V. BCLEARER'S BROAD STRUCTURE

The bCLEARer process has been modularized (see [53, App. B], [54]) into a component architectural pattern. We describe this in the first section.

In the earlier comparison of current methodologies, we assessed them relative to two levels: generalization and digitalization. We now describe how bCLEARer addresses these levels in the second and third sections below. In the final, fourth, section we look at whether it is better to surface-*computerize (using bCLEARer) *in vitro or in vivo*.

*1) bCLEARer's Pipeline Component Architecture Framework*

The bCLEARer process has a pipeline (pipe-and-filter) architecture [54], a prevalent approach for data transformation. This architecture consists of a sequence of processing components, arranged so that the output of each component is the input of the next one creating a 'flow'. The pipeline architecture has, as the 'pipe-and-filter' name suggests, a series of pipe and filter components, where pipes pass data to and from filters that transform the data — the pipeline flow. The architecture can be nested, in that filters can encapsulate a sub-pipeline process.

This generic architectural pattern is refined into a more constrained pattern for bCLEARer's more specific needs. It must include the components of the ontologization process in a structure where the specific arrangement of components can be dictated by the needs of the project and this arrangement can flexibly evolve over time, potentially into a radically different shape.

Typically, it is divided into three broad levels:

1. thin slices – which typically correspond to ways of dividing the domain and the dataset [55]
2. bCLEARer stages – the stages that correspond to a particular type of transformation

3. bUnits level – the filters within a single bCLEARer stage, the base filters are called bUnits.

*a) The bCLEARer stage types*

While the contents of the individual thin slices and bUnits level vary from project to project depending upon their needs, as well as evolving over time, the bCLEARer stage types are a more stable architectural feature. The design of these types is motivated by the 'separation of concerns' [21] principle – where each type deals with a different kind of transformation. This builds upon the factorization discussed above. The five stage types are Collect, Load, Evolve, Assimilate and Reuse (whose initials contribute to the acronym bCLEARer).

Collect is the stage at which a dataset enters the pipeline. Collect stores the dataset and ensures it is not changed. There is no transformation at this stage. This provides a fixed baseline for tracking. Larger datasets are divided into chunks, to be consumed one chunk at a time.

The Load stage receives the dataset from the Collect stage. The first thing it does is establish the identity of the contents to facilitate tracking and tracing. The Load stage is responsible for ensuring that the data passed onto the next Evolve stage is *computerized – at least surface-*computerized. If the dataset comes from an operational application system that uses an enterprise database, the data will probably be sufficiently structured and so need no *computerization transformation. If it is unstructured text, for example a PDF text document, it will be pre-*computerized, and so need transforming. The Load stage undertakes the minimal amount of transformation to *computerize it, in effect it surface-*computerizes it. Where this is required, the project will need to decide on the output format to use. In our projects, we usually make the target structure simple tables.

The Evolve stage assumes its input data is (at least) surface-*computerized. It is responsible for digitalizing this input data. This is done in two major sub-stages. First it deep-*computerizes the data and then *ontologizes it. Typically, the very first exercise in the deep-*computerize stage is to check whether the data needs cleaning, and if so, clean it. When the data comes from several systems, it normally makes sense as part of the deep-*computerization stage to integrate the data across systems into a common format, as far as possible, after firstly transforming the data from each system on its own. When the deep-*computerization is complete, the *ontologization can start. This is guided by a minimal foundation, the BORO Seed – for an example of a relatively recent minimal seed see *Top-Level Categories* [56]. A full digitalization project will include both *computerization and *ontologization. But pragmatic considerations may dictate that this is done in phases – and the early phases may only go so far along the digitalization journey. For example, undertaking deep-*computerization and delaying *ontologization to a later stage.

The Assimilate stage assumes its input data is 'evolved' – so both locally *computerized and *ontologized. It is responsible for assimilating this into a common cross-project model. The assimilated model is then ready for use in future Assimilate stages.

The Reuse stage assumes its input data is assimilated. It is responsible for translating this data back into a format usable by the targeted operational systems.

*b) Managing micro-coevolution*

To some extent, the discussions about factorization and components shift focus away from the micro-coevolution that takes place. The bCLEARer journey typically involves evolutionary adaptations simultaneously on two fronts:

- Information Evolution: Adaptation of information throughout its journey.
- Journey Evolution: Adaptation of the journey itself to emerging requirements, accelerating the information's evolution.

The whole process supports both these adaptations: identifying and accommodating significant changes in both the information and its digital journey. A key element is adaptive resilience: maintaining stability and efficiency of the factorization and components amidst continuous change.

*2) A bCLEARer example*

A concrete example of how *computerization and *ontologization are deployed in the first three bCLEARer stages might help to make some of these points clearer. Let us say we have a legacy migration project that encompasses intercompany accounting systems. We have three source systems: PHAS (Peak Holdings Accounting System) from Peak Holdings Ltd, AAS (Acme Accounting System) from Acme Ltd and ZAS (Zenith Accounting System) from Zenith Inc., where the latter two companies are owned by Peak Holdings. For simplicity, assume all three systems are being migrated to a new COTS system – NAS (New Accounting System). Assume there is also a desire to take this opportunity to harmonize the accounting practices across the three systems. Using bCLEARer provides a systematic approach to not only integrating the data but also exposing digitalization opportunities.

*a) Collect stage*

The Collect stage will initially take a snapshot of the full dataset from the systems for the initial development of the pipeline. Later snapshots will be taken as required. As these datasets come from operational systems, they have a holistic coherence and consistency that we need to make sure persists through the bCLEARer process.

*a) Load stage*

For simplicity, assume the PHAS, AAS and ZAS systems all use relational databases. The datasets are already surface-*computerized so there are only a few further specific *computerization tasks needed at this stage. The first task in bCLEARer Load is always to mark the identity of the data, to provide a baseline for tracking and tracing. We give the systems, tables, columns and rows identities – and, where necessary, the cells as well. This is typically done with a cryptographic hash function.

We also identify and inherit from the source systems the queries that can be used to check for coherence and consistency. These would include standard reports such as, in this case, the account ledger, balance sheet and profit and loss reports. We typically run and hash the figures in the reports so we can easily run a simple automated binary comparison check.

*a) Early Evolve stage – *deep-computerization*

The early Evolve *deep-computerization stage is approached with an ethnographic mindset, interpreting and understanding the dataset's implicit structure from its own perspective – aiming not to introduce any biases. This opens the possibility for a multiplicity of syntactic changes (adaptations) – data cleansing being one type.

*b) Early Evolve design pattern – unification of types*

The Evolve stage focuses on deep-*computerized. We have developed a range of design patterns to facilitate this stage. One useful design pattern simplifies the handling of data formats. There is no restriction of the format in which the dataset comes in at the Collect stage. It could be in XML, JSON or SQL or a combination of these or other formats.

However, for the ontologization process these specific implementation data formats are an irrelevancy so can be filtered out. The higher levels of (syntactic) generality, the metadata and schema can be mapped into the data, which removes the dependency on any specific implemented format. We call this mapping the unification of types [37], [57]. This enables us to choose for this stretch of the pipeline a data format that suits the work we want to do – and build common code for this. It also greatly simplifies making the metadata explicit – as what was built implicitly into the collect data form can now be made explicit.

In the case of the three systems, which use relational databases, the tables and columns are shifted into the data – unifying the schema and the data. At the same time, the balance sheet, profit and loss and other queries are adapted to the unified data structures and used to test the relevant semantics are preserved. There is no gap between the migration to the unified structures and testing with the queries.

When we have unified the types, we have standard tools to graph-visualize the data. We do this at each major stage along the pipeline. We have found (and it is well-recognized) that it is a good way to handle large quantities of data.

*a) Early Evolve stage – syntactic integration*

Typically, different systems implement what is clearly the same information in different structures, sometimes very different structures. This creates opportunities for syntactic integration. For example, the format for the chart of accounts and postings is likely to vary between the three systems. At this stage, we take the opportunity to make simple changes that harmonize the information in the three systems, taking care to respect the perspective of the individual systems.

*a) Later Evolve design pattern – *ontologization*

Once the opportunities for deep*computerization have been exhausted, if appropriate we move to the later Evolve stage and start the *ontologization. However, there may well be situations where it makes sense to delay this until some future project.

*Ontologization typically involves making semantic adaptations. We have *ontologized accounting systems before and seen the kind of semantic adaptations that emerge, see [58], [59], [60], [61], [62]. One adaptation these identify – see [62] – is the shift from *de se* perspectival accounting to *de re* 'objective' accounting. We would expect this adaptation to emerge here as well.

The way it would emerge is as follows. The top ontology would provide criteria for identity. When these are applied to individual intercompany transactions, this will provide the basis for recognizing where transaction and accounts are the same. However, under current accounting conventions these will be marked with opposing debit and credit properties. Transactions and account balances that are marked in one system as debits will be marked as credit in the other system. It turns out that whether these are tagged as debit or credit is subjective and depends upon which company's perspective is taken. One then recognizes debit and credit as a relational property between the transaction and the company.

In more practical terms, it means that the form of the data is changed. All the identical original accounts and transactions are merged into new ones – and a debit/credit relation between the company and them are established. These changes start in the data and are propagated into the schema. At the same time, the queries are amended (evolved) to take account of the new structure – and tested to ensure they can reproduce the figures in the original reports. They both confirm the consistency of the new structure and help to ensure the adaptation is preserved along the pipeline. One can recognize this as an empirical exercise where the changes emerge from the data. It is hard to see how rational inspection of the schema without consideration of the data could lead to this adaptation.

*3) bCLEARer's level of generality approach*

We introduced the broad division of information from a data perspective into these levels of generality: the metadata, schema and data levels. We can use this to classify the methodologies' information pathways and so clearly differentiate the methodologies.

*a) Scoping level of generality*

We can see differences in the 'generality' scope of the methodologies' information pathways. As shown in Fig. 2, the mainstream 'ask-an-expert' (AaE) and 'top-down-classification' (TDC) methodologies differ in that AaE scopes the metadata level out and TDC scopes it in. The two mainstream methodologies differ from bCLEARer in their scoping of data. The two mainstream methodologies scope the data level out and TDC scopes it in.

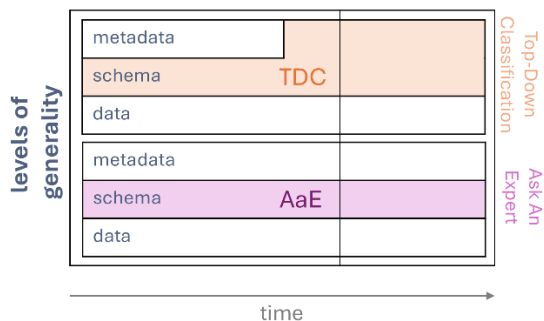

Fig. 2. Methodology information pathway – levels of generality

This descoping is reflected in the standard ISO 21838-1 where a domain ontology is defined in section 3.18 thus: "ontology (3.14) whose terms (3.7) represent classes (3.2) or types and, optionally, certain particulars (3.3) (called distinguished individuals') in some domain (3.17)" – and later says "Some ontologies also allow terms representing certain privileged particulars (referred to as 'distinguished individuals'), such as 'the actual world', 'spacetime', or (in an ontology of US law) 'the US Supreme Court'". The standard recognizes that the domain includes particulars, as it is defined in section 3.17 thus: "collection of entities (3.1) of interest to a certain community or discipline" with a note to say "'Entities of interest' can include both particulars and classes or types." The standard assumes (without explanation) that unless something is a special 'distinguished' particular, it is excluded from domain ontologies.

This radical descoping approach would make sense if one somehow could acquire a high degree of confidence that the rationally designed schema structures would adequately support the requirements of the domain data. In software engineering, it is well-recognized that we do not have the tools to provide this confidence, that one needs to empirically test the schema with data to acquire a sufficient degree of confidence. So, the radical descoping of data effectively eliminates any direct testing of schema patterns that involve data – in other words empirical testing.

Historically, projects have typically built in a delay between the rational design and the empirical testing. One can see this clearly in waterfall-stye software development, where significant data has historically only been introduced into the project for volume testing.

However, the more one delays including sufficient significant data in scope, the greater the gap between the introduction of a defect and the possibility of finding and fixing it and the more difficult and expensive it becomes to fix. Hence the development of methodologies such as Extreme Programming [63] that aim to identify defects early. And the emergence of discussion of a shift left approach [64], where testing is performed earlier in the lifecycle. More recently, DevOps has embraced this approach.

Delaying the testing may once have been a justified pragmatic choice. Typically, in most systems there is far more data than schema – and more schema than metadata. So, an initial descoping of data will usually have the effect of significantly reducing the size of the dataset, making it feasible to manage with existing technology. In the last decades of the 20th century, there may have been good technological reasons for working this way, but this is no longer the case.

The bCLEARer approach is fundamentally empirical. It expects the design patterns to emerge from the data. Design patterns are described in terms of the data that exemplifies them. Hence there is literally no gap between design and testing – they are the same process.

### b) Top-level ontology deployment

There is an element of ambiguity about the metadata level that becomes clear when we consider bCLEARer. In the 'top-down-classification' and 'ask-an-expert' approaches the main information pathway starts with unstructured data, which has no top-level metadata (though it may have provenance 'metadata'). In the 'top-down-classification' approach the top-level metadata is a previously prepared top-level ontology.

When bCLEARer is processing a structured dataset, this will have top-level metadata – the structure of the data into, for example, tables, columns and rows. During the deep-*computerization stage, this will be made explicit, where it is not already. At this stage one works with the data rather like an ethnographer working within a culture – aiming to make its implicit, invisible assumptions explicit without imposing one's own views, especially on the nature of the domain.

At the *ontologization stage, the situation is different. The ontological commitments are usually far from clear and often the top-level commitments completely inscrutable. bCLEARer gets around this by introducing a top-level, but it does not want to let this interfere with the underlying picture of the domain. So, bCLEARer aims for a balance where the top-level is sufficiently ontologically rich and complex to guide the analysis effectively, but also sufficiently minimal that it does not hinder, or block refinements emerging from the bottom (data) or otherwise render the validation ineffective. One aims to seed the process with a top-level that is sufficient to make the ontological foundations, and so the ontological commitments, scrutable. This could then guide the *ontologization. One also aims to make this as minimal as possible to maximize the benefits of bottom-up grounding. To be as open as possible to refinement as the lower-level ontological commitments emerge from and are confirmed in the data. For a discussion of how to construct a minimal seed see *Top-Level Categories* [56].

### 4) Navigating levels of digitalization – the information pathway

As the two methodologies we looked at earlier show, an ontologization process will have an information pathway that navigates the levels of digitalization. How should one design this navigation?

### a) Shifting digitalization right or left

One can characterize this pathway in terms of whether the transformation to *computerization (from unstructured to structured data) is performed earlier or later in the lifecycle (that is moved left or right on the project timeline). Fig. 3 shows us this for the three methodologies we are looking at.

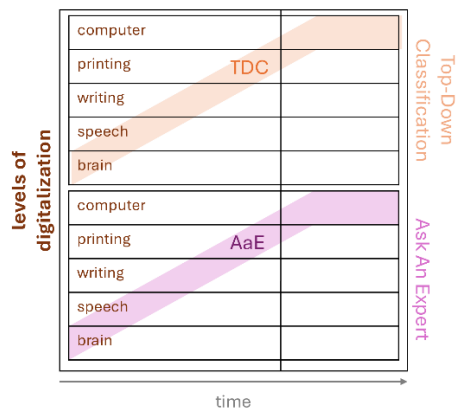

Fig. 3. Methodologies – Levels of digitalization

The 'ask-an-expert' (AaE) and 'top-down-classification' (TDC) approaches, looked at earlier, shift right. They move the transformation to *computerization towards the end of the project lifecycle (and *ontologize at the same time, without intermediate steps).

Data pipeline approaches, such as bCLEARer, shift left. They aim to make the information pathway transformation to *computerization as early as possible. The ideal case is when the collected dataset is already structured so already *computerized.

What motivates this position is the aim to be in a situation where the data pipeline can run automatically, getting into 'machines talking to machines' territory as soon as possible. One only needs surface-*computability, not deep-*computability. Hence, one gets all the benefits of automation as early as possible.

*b) Example – definitions as a marker of digitalization maturity – shifted right or left*

A good marker of the level of digitalization of an approach is how it handles natural language definitions for humans. The level of maturity of this process is a good guide to the overall level of digitalization.

[3] is a good example of a common practice. It has 8 pages (pp. 68-76) and the same number of principles (13-20) devoted to how to write (in natural language) definitions properly. The first principle (13) states: "Provide all nonroot terms with definitions."

From an information pathway perspective, the information is assembled in a human brain and translated into natural language text. Then it is (hopefully) used as an input to a later manual formalization: as the text is not computer readable. Also (hopefully) when the formalization changes, then effort is made to manually bring the natural language text in line. This effort is manual, so is prone to error and does not scale. Clearly, this process is at a pre-*computerization level of digitalization.

Stafford Beer [65] said, "the purpose of a system is what it does" (converted into the acronym POSIWID) and, to drive the point home it has been observed that there is no point in claiming that the purpose of a system is to do what it constantly fails to do. The same point applies, more narrowly, to a computer system's use of a term. If the system (somehow) uses the term in a certain way, then that is surely what the term 'means' to the system. If one writes the code for the system in a 'clean' way [50], we can algorithmically translate the way it uses the term from the machine language of the system into something human readable. Surely this 'definition' with its direct connection with what the system actually does is far more trustworthy than something produced by humans on behalf of the system. And we can also trust that as the system evolves and changes, this 'definition' will change with it.

Hence, bCLEARer-type approaches aim to clean the data in the pipeline so that the natural language definitions can be algorithmically extracted directly. From an information pathway perspective, all the definitional work is automated, predicated upon at least cleaning the data. This avoids all the manual effort that would have been devoted to these definitions. Clearly, this process is at least at a *computerization level of digitalization, and at an *ontologization level if one needs better output.

*5) Digitalization – surface-*computerization – in vivo versus in vitro*

Directed evolution introduces an *in vitro* mimicry of *in vivo* evolution. This raises a natural question about what factors affect the choice between these two approaches. Obviously, a big factor is how successfully one can target good adaptations.

Given that we have operational enterprise systems, then it is plain that *in vivo* evolution can produce working computer systems (so surface-*computerized data). There is also lots of evidence of failed projects, so we know that it is not easy.

When we have these systems, the problems of dirty data are well known. If one looks, in any detail, at the data innards of successful enterprise operational systems it is surprising how well they work given how 'dirty' and disorganized they are. Data cleansing exercises show how relatively easy it is to improve them. This suggests that one can target good adaptations for the deep-*computerization process. bCLEARer's experience agrees with this.

When one works with these systems, it soon becomes clear that most have little or no clear ontological commitment. One simple test is to see how the system handles mereology – it is rare to find a system that has a clear picture of this. Similarly for multi-level types [22]. bCLEARer's experience suggests that if done properly then building in a clear ontological commitment is feasible and can reap significant benefits.

Together these suggest that one can target good adaptations for deep-*computerization and *ontologization provided one has the surface-*computerized data. It also suggests that *in vivo* natural selection is not so good at finding these adaptations, at least in the timescale these systems have had to evolve.

From a more general perspective, this suggests that the role of ontologization is not to facilitate the first step in digitalization – surface-*computerization, but more to enable a second step that significantly improves systems.

*a) Experienced systems*

It is normal to think of computer system quality degrading over time, that as systems get older, they accumulate technical debt and the quality of their data declines. It may well be true that, as the amount of data in a system grows, the amount of erroneous data items also grows. But this misses an important point from our perspective, that systems accumulate a kind of experience over time. Typically, both the variety and complexity of their data increases and it becomes a better reflection of the domain. This reflects the enormous investment in both the operation and maintenance of the system.

From the bCLEARer-type data pipeline perspective this 'experience' is valuable. As Ashby's discussion [66] of variety makes clear the richness and complexity of the picture we build of the domain will depend upon the richness and complexity (the variety) of the data we use to build it.

From one perspective, this is a classical evolutionary situation. A biological unit (in this case, a computer system) garners information about its world that is useful to it. Unless this information is heritable, and inherited, then it stops being useful when the unit dies. Genetic inheritance is a very lossy way of transmitting information. Pipelines like bCLEARer offer the prospect of salvaging significant portions of the data.

From another, breeding perspective, it suggests a rule of thumb. Given that we aim to harvest domain patterns from the digitalization exercise, then if we select more experienced operational systems – and several of them – then we will harvest richer more accurate patterns.

*b) Harvesting pre-\*computerization*

There is a flip side to this. What should we do in cases where there is not even an operational system, let alone an experienced one. We can deploy ontologization processes on pre-existing unstructured data or even synthesize unstructured data. The synthesized data will not have been subjected to any real selection pressures. The unstructured data may have been subject to some selection pressures, but these will not shape its formal structure (as it is unstructured). In these cases, there seems to be a lack of variety, or at least the right kind of variety.

A rationalist might think armchair reflection will be able to provide requisite variety. But this will be rooted in brain wetware, speech and writing – all legacy technologies from a computing perspective. Is this good enough to target good, computerized adaptations? There is a lack of data on this topic, but our anecdotal evidence is that it falls well short of what is required. That without the computer experience to build upon, our targeting falls back on legacy technology patterns of thought that prove unsuitable for the scale and formal precision of computerization.

VI. SITUATING DATA PIPELINES AS EVOLUTION

One can see the goal of a bCLEARer-type data pipeline project is to build an *in vitro* high evolvability environment for information, providing it with the possibility of evolving fast. This environment is easier to build when one develops a sensitivity to the complex set of drivers that the evolutionary perspective reveals.

The evolutionary perspective is an incredibly rich resource, and we are still in the process of understanding how the digitalization process fits into it. However, there are several elements of the perspective that we have found useful, and we outline a few of these in this section to provide a sense of what the perspective entails.

*1) Data pipelines as directed or experimental evolution*

From an evolutionary perspective, the bCLEARer-type data pipeline methodology can be seen as a kind of directed [67] (or experimental [68]) evolution – where experimental evolution is sometimes called "laboratory natural selection".

Direct evolution is used in protein engineering – and contrasted with rational design, which targets specific point mutations. However, from a broader perspective, the range goes from random natural mutation to deterministic rational design with directed evolution somewhere in the middle. But the rate of natural mutation is usually insufficient for generating the genetic diversity required for laboratory directed evolution, so it is out of scope. Directed evolution balances the difficulty of 'rationally' accurately predicting how specific mutations will impact protein function with the slow and unpredictable pace of natural selection. It uses targeting but reduces the need for accurate predictions replacing by iteratively selecting mutations and 'empirically' testing them.

In the digitalization context, rational design would be attempting to build the ontologization from first principles, whereas the directed evolution would start with the existing data and target a range of likely mutations and iterative tests which lead to better adaptations. In a process broadly analogous to directed and experimental evolution, it repeatedly targets and constructs variants in an iterative step-by-step process, continually inspecting the results and selecting for fitness, aiming to mimic natural selection.

This evolution is directed in the sense that there is an element of Lamarckian target setting when managing the mutations, which, when successful, speeds up the evolution. The direction cultivates and nurtures evolution – focusing on fostering and guiding innovation – rather than purely analyzing or breaking down data. The aim is to guide the selected dataset (of information) along a journey of digitalization transformation that exploits the opportunities offered by digital technology. More specifically, to exploit the opportunities for *computerization and *ontologization of information.

*2) Data pipelines as evolving information transmission*

Inheritance involves the transmission of information between individuals. Genetic heredity involves passing information from one generation to the next – often called vertical transmission. Sexual reproduction is an example of vertical transmission – transmission from parents to their offspring.

If, as suggested earlier, we see computer systems as biological individuals, then within digitalization, legacy system replacement can be seen as a good example of vertical transmission, where the data (information) in the legacy system is transmitted to the new system.

A bCLEARer-type data pipeline works at the level of information transmission – in other words, transmission between systems. If deployed in the legacy system replacement case, it would take control of the transmission of information from the legacy to the new system. So, the pipeline can be seen as an example of digitalization information transmission, with a focus on developing adaptations during transmission between computers.

Developing adaptations during information transmission is not new to evolution. In natural evolution mechanisms for creating adaptations during information transmission have arisen, sexual reproduction being a classic example. The genetic recombination of genetic material from two parents can introduce novel variation.

bCLEARer-type data pipelines are designed for fast evolution. For good empirical reasons, biological experimental evolution, mentioned above, will often adopt a life span speed strategy. It will select individuals, such as fruit flies, with a short life span, to enable testing to occur over multiple generations and so speed up evolution. However, this strategy of shortening life spans and increasing the number of generations makes less sense in the pipeline case, where (among other things) there is not a plethora of systems with short lifespans. However, the general strategy of increasing the pace of evolution stands. pipeline achieves this through both extending and enriching the information transmission process as well as iterating it – mimicking the evolution of multiple generations within a single transmission.

*3) Challenge – uncertainty, contingency and chance*

The uncertainties around innovation are well-known [69]. One way these uncertainties are framed in evolution is as contingency [70], [71], This recognizes that evolution is a historical process and so is sensitive to and so contingent upon, the paths taken in its journey – in other words, sensitive to chance.

In general, the potential for innovation is usually so wide-ranging, so subject to chance that it makes no real sense to ask whether an opportunity has been missed. However, in the restricted digitalization context, we are designing systematic processes that more regularly lead to innovations and so making adaptations significantly less subject chance. In this situation, it makes sense to ask whether we are missing innovation opportunities that we could (should?) have spotted.

*a) Macro- and micro-evolutionary contingency*

To illustrate evolutionary contingency, Stephen Jay Gould [72] used the thought experiment of rewinding the "tape of life" to the distant past. He argued that even small changes to the path of history could result in evolutionary outcomes very different from our world, such as, for example, no humanity.

For our purposes, we can usefully distinguish between broadly global macro-contingency and local micro-contingency. Where global macro-contingency is whether a particular major outcome will ever (globally) happen – for example, humanity or language or computers emerging. And local micro-contingency is whether a particular minor outcome that could happen will do so in a local situation. We see micro-evolutionary contingency when different similar beetle populations respond differently to the same pressures – such as climate change.

*b) Evolutionary data pipeline contingency*

We can translate macro- and micro-evolutionary challenges to our bCLEARer-type data pipeline digitalization context.

At the macro-evolutionary level, we recognize that it is not inevitable that humanity will exploit the major opportunities of digital technology. We have already noted that the exploitation of technology depends upon the appropriate co-evolution of technology and cultural practices. And that the evolution of the cultural practices depends, at least in part, upon the appropriate Lamarckian targeting. If this doesn't happen, the innovation opportunity will be missed. The (broadly) global question asks whether it will happen in our (near) future.

The concern is not entirely theoretical as we have examples from the past. Olson [33] also describes how Western European culture successfully evolved to take advantage of printing technology when cultures in other parts of the globe (such as China) did not, even though they had earlier access to the technology. This provides us with a good illustration that technological innovations need cultural variations that will successfully exploit selection pressures, that these evolutionary pathways are contingent upon taking (targeting and construction) a potentially successful direction.

Pragmatically, contingency concerns are about completeness – about how exhaustive the process is. At the macro level, the goal is to design a framework whose use is likely to maximize the chances of finding and exploiting the general opportunities, particularly the most fruitful opportunities, for *computerized and *ontologized digitalization. At the micro level, the goal is to design a pipeline using the framework whose operation is likely to maximize the chances of finding and exploiting the specific opportunities. At both levels, the aim is to minimize the risk of overlooking valuable opportunities.

bCLEARer is an example of such a framework at both the macro- and micro-levels. It is designed as a tool to systematically find and exploit opportunities for *computerized and *ontologized digitalization.

*4) Data pipelines as in vitro evolution*

If one restricts one's perspective to bCLEARer-type data pipelines, then most of the process is (in a sense) *in vitro* – in a walled garden outside the original system. However, if one steps back the pipeline usually plays a role in a wider live *in vivo* ecosystem.

Also, there are usually important (in a sense) *in vivo* tests where the information is returned to at least one operational system and tested 'in the wild'. If possible, this is to both the original and similar systems. In an ideal configuration of the pipeline, the improvements are fed back into the original system on an ongoing basis and the results inspected.

VII. OUTLIER DESIGN CHOICES

In bCLEARer-type data pipeline information transmission there are two evolutionary processes each with their own information pathway. There is the information being processed by the pipeline and then there is the pipeline process itself – as code. Both are co-evolving intertwined in a process of reciprocal causation – each feeding of the other.

The information being processed by the pipeline is the full data set, with no level of generality being excluded. The pipeline process is as automated as possible, so its information pathway as *computerized as possible – in other words, is digitalization shifted left as far as possible. Both these are outlier design choices. In this section we look at ways of explaining aspects of these outlier choices.

*1) Should transmission include data inheritance*

There is a further refinement of the evolving information transmission narrative related to the role data plays in it. In the methodologies we have been looking at, under the level of generality accessibility perspective, one can choose whether to include data (the lowest level of generality) in the ontologization process. Simplifying a little, the 'ask-an-expert' and 'top-down-classification' methodologies exclude data, a bCLEARer-type data pipeline methodology includes it from the start. (The simplification is that the metadata-data-schema classification is about the syntax of the implementation, whereas generality is a semantic matter. However, there is good enough rough match between syntax and semantics here to make the point fair.) This is an all or nothing choice. Unlike in software development methodologies, especially waterfall, where the full dataset is included in the process part of the way through – typically towards the end. Hence, we label this as an architectural design choice on whether to shift (far) left or (far) right.

### a) Weismann's distinction

One can get a sense of this architectural design question from a distinction make in 19th century evolution theory. Weismann [73] turned the point that the mechanisms of transmission typically can only transmit some information in the source into a distinction. He made a basic (since refined) division of cells into the germline and the somatic line which gives us a neat, simplified picture of the underlying structure. These are similar to the *The Selfish Gene*'s [36] "replicators" and "vehicles".

The germline is those cells that are involved in reproduction and the transmission of genetic information from one generation to the next. Mutations in the germline are crucial for evolution because they can be passed to the next generation, in other words, they are heritable.

The somatic line is the rest of the cells, the non-reproductive cells. They are not involved in the same way in transmission. Mutations in these cells may affect the individual and so their fitness, but are not transmitted on to offspring, so they are non-hereditary.

The germline cells have been called 'immortal' in the sense that they (or their genes) continue to exist indefinitely through reproduction – creating a lineage. Whereas the individuals and their somatic line cells die, they are mortal. Thus, changes in the germline can contribute to genetic diversity and evolutionary adaptation, while changes in somatic cells affect only the individual organism's health or fitness.

If one maps Weismann into the world of computer systems, then there is a recurring pattern of transmissions where the data-schema division aligns with the germ-somatic line division, where data is heritable, and schema is not.

One clear example is the migrations between COTS systems which have an analogous structure to vertical genetic transmission. The data is migrated (transmitted) from the old application to the new application – and so is immortal in the sense it persists between generations. The schema (and the rest of the application) is like the somatic line in that it 'dies' with the old application. APIs (Application Programming Interfaces) also have an analogous transmission structure. The data is transmitted between applications whereas the schema is not.

This data-immortal, schema-mortal picture is, like Weismann's, a simplification. But it is broadly true in that the data persists much longer than the schema – though as it moves between applications it gets mapped to the new applications schema.

We can frame this in economic terms. If we think of a biological unit (whether organic or silicon computer application) as storing information as an investment, one which 'pays' a return when used. Then information transmission can be seen as a way of preserving that investment across units to generate better 'returns. When this insight is combined with the realization that in many current transmissions data is transmitted and schema is not, then data would seem to be a better place to invest in the information system ecosystem.

### 2) Data as embodied competencies

The use of competency questions is a rationalist approach. At its simplest, it assumes that we have sufficient knowledge to unaided target competencies that we require and then construct a computer system with the competencies.

bCLEARer-type data pipelines are examples of an empirical approach. They start with source operational systems that we can verify have a certain level of competence. The datasets from these systems embody these competences. They must do, otherwise the systems would not operate. One can make these competencies explicit, exhibit them, through queries on the datasets – ones which are often already built into the source systems.

Over time, the data structures in computer systems are twisted and turned to accommodate new requirements. Hence, there is an understandable feeling that the structures are somehow defiled, unclean. While it is probably true that the lack of cleanliness holds back some level of competency, it is not true that it indicates a (total) lack of competency. The computer systems operate, often at a sophisticated level, they still have the competencies. The *ontologization stage of the digitalization process addresses this lack of cleanliness providing a hyper hygienic level of cleanliness that lets new competencies emerge.

### 3) Managing the inheritance – preserving and improving the investment

If one shifts right and includes data in the process, then one is faced with a responsibility for managing that data.

### a) Transmission fidelity and transformations

Transmission fidelity ensures that the transmitted information maintains its original shape and characteristics throughout the transmission. In genetic (DNA) inheritance a reasonably high transmission fidelity is needed to maintain organismal stability across the generations. But mutations, a failure of fidelity, are the variations that provide the raw material for evolution. So, if we want adaptation and natural selection to occur we need to ensure we have mutation and so variation.

In the pipeline, the formal nature of digital computing means fidelity works in a different way. Though there is some degradation of the digital signal in some circumstances, this is not significant. So, we can pragmatically assume digital fidelity. We still need new variations, but these are formally created by the pipeline code.

DevOps recognizes the importance of pipeline observability engineering [74]. The term is borrowed from control theory, where the "observability" of a system measures how well its state can be determined from its outputs. Majors et al. [74] suggests that what differentiates observability is its focus on not just identifying issues but aiming to minimize the amount of prior knowledge needed to resolve an issue. This has, historically, been a significant driver for bCLEARer where significant time used to be lost attempting to track and trace adaptations along the information pathway. Where tracking follows information, and tracing follows how information has influenced other information. This has led to the bCLEARer pipeline introducing an additional kind of observability – what we call 'inspectability' – which is the ability to map in full detail the transformation journey along the information pathway.

We have been working over the last few decades evolving an inspectability framework. The specific goal of this inspectability framework is to be able to track and trace the items of information through the pipeline. This relies firstly on having a clear notion of identity for these items. This means, ironically, we needed to build an ontology for the information in the pipeline. We need to be able to extend this ontology to give us a clear notion of tracing – relating how items are transformed into new items. We then needed to build infrastructure into the pipeline to make this ontology explicit. Finally, we needed to able to access this ontology at regular inspection gates and have tools that allow us to view and visualize it.

With this in place we can track and trace information items and their transformations through the pipeline, between pipeline runs and between pipeline evolutions. One useful visualization is the information items' ontogenic tree – analogous to the phylogenetic tree – showing how the information items transform as they pass along the information pathway, as well as how data and schema coevolve.

*b) Automation and the dataset*

Automation has improved the pace and scale of digitalization's directed evolution. It is well known that pace is a key factor in being able to generate change in a reasonable time. In evolutionary research the fruit fly has a key role due to a very short life cycle, typically around 10 days from egg to adult, leading to fast evolution. In innovation research, Christensen (see *The Innovator's Dilemma* [69]) picked the disk drive industry because of its fast pace of change, referring to it as the 'fruit fly' of the business world.

Pace is similarly important in ontologization pipelines. It has a couple of aspects which we have already noted a few times. The first and simplest is the pace of a single pipeline run – the information evolution. This needs to be quick enough to allow for frequent runs. The second is the pace of the evolution of the transformations in the run in a project – the project process evolution. The third is the pace of the evolution of the transformations across projects – the process evolution.

A major impact on pace, as well as the investment required, is the development of the pipeline code. An important way to reduce costs was to evolve common code, where code is reused rather than written anew from scratch. The aim is for much of the final code to be common to multiple bCLEARer data pipeline projects. There are opportunities to build common code for the running of the pipeline. There are also opportunities to evolve general patterns of transformation (and the components of the transformations). One can see this as digitalizing the transformations – where the transformations are carried out by machines on machines. Using machines to build better machines has a long history. One well-known episode is the use of John "Iron-Mad" Wilkinson's machine to precisely bore the large cylinders needed for James Watt's steam engines – significantly improving efficiency over the previous manually crafted ones.

Achieving the goal of a common codebase requires the adoption and coordination of multiple techniques. There are a variety of software development hygienes that reduce the cost of maintenance and enhancement such as clean coding [50]. There is also the continuing evolution of design patterns facilitated by a close analysis of the transformations.

VIII. CONCLUSION

We have used a two-dimensional analysis (over generality and digitalization) to identify new data pipeline-based opportunities for exploitation in the ontologization methodology design space that are not exploited by current mainstream. From a levels of generality perspective, there are opportunities to be more inclusive with data from the start of the pipeline. From a levels of digitalization perspective, there are opportunities to shift the computerization of the process to the far left, to the start of the pipeline. We have used bCLEARer as an example of how this can be done.

We have noted that engineering of the ontologization process is design poor and raised the need to remedy this. As part of this remedy, we have factored the process into separate concerns. Divided it firstly into *computerization and *ontologization and then further divided *computerization into surface-*computerization and deep-*computerization. We have suggested that we should consciously design the order of these processes.

Finally, we have called attention to the point made in many fields, that setting the right context is critical to success. In this field, we suggest that a fruitful context is information evolution. We describe how this evolutionary perspective situates digitalization as the latest iteration in the overall evolution of information transmission. And then situates *computerization and *ontologization as key cultural practices in digitalization's coevolution.

ACKNOWLEDGMENT

We wish to thank Mesbah Khan and Andreas Cola for their helpful reviews of the paper.